# The Basque task: did systems perform in the upperbound?

**Eneko Agirre, Elena Garcia, Mikel Lersundi, David Martinez, Eli Pociello**
IxA NLP group, Basque Country University
649 pk.
20.080 Donostia, Spain
eneko@si.ehu.es

**Abstract**

In this paper we describe the Senseval 2 Basque lexical-sample task. The task comprised 40 words (15 nouns, 15 verbs and 10 adjectives) selected from *Euskal Hiztegia*, the main Basque dictionary. Most examples were taken from the *Egunkaria* newspaper. The method used to hand-tag the examples produced low inter-tagger agreement (75%) before arbitration. The four competing systems attained results well above the most frequent baseline and the best system scored 75% precision at 100% coverage. The paper includes an analysis of the tagging procedure used, as well as the performance of the competing systems. In particular, we argue that inter-tagger agreement is not a real upperbound for the Basque WSD task.

## 1 Introduction

This paper reviews the design of the lexical-sample task for Basque. The following steps were taken in order to build the hand-tagged corpus:
1. set the exercise
    a. choose sense inventory
    b. choose target corpus
    c. choose target words
    d. select examples from the corpus
2. hand-tagging
    a. define procedure
    b. tag
    c. analysis of inter-tagger agreement
    d. arbitration

The following section presents the setting of the exercise. Section 3 reviews the hand-tagging, and section 4 the results of the participant systems. Section 5 discusses the design of the task, as well as the results, and section 6 presents some future work.

## 2 Setting of the exercise

In this section we present the setting of the Basque lexical-sample exercise.

### 2.1 Basque

Basque is an agglutinative language, that is, for the formation of words, the dictionary entry independently takes each of the elements necessary for the different functions (syntactic case included). More specifically, the affixes corresponding to the determinant, number and declension case are taken in this order and independently of each other (deep morphological structure). One of the main characteristics of Basque is its declension system with numerous cases, which differentiates it from the languages spoken in the surrounding countries. An example follows (the order of the lemmas is the reverse):

etxe<u>koari</u> emaiozu
[Give it] [<u>to the one in the</u> house]

### 2.2 Sense inventory

We chose a published dictionary, *Euskal Hiztegia* (Sarasola, 1996), for the sense inventory. It is a monolingual dictionary of Basque. It is normative and repository of standard Basque. It was produced based mainly on literary tradition. The dictionary has 30,715 entries and 41,699 main senses (see comment on *nuances* below). The TEI version with all the information for each entry was included in the distribution. As the format was quite complex, another version was also included, which listed a plain list of word senses and multiword terms.

This dictionary has the particularity that word senses can have very specific sub-senses, called *nuances* which sometimes are illustrated with just an example and other times have a full definition. These *nuances* were also included in the set of word senses.

### 2.3 Corpora used

At first the EEBS balanced corpus was chosen, comprising one million words. Unfortunately this size is too small to provide the number of occurrences per word that was defined in the Senseval task specification. We therefore turned to the biggest corpus at hand, the *Egunkaria* corpus, comprising texts taken from the newspaper. The size of this corpus allowed us to easily reach the number of examples required. On the negative side, it is a specific corpus, and the distribution of the word senses could be highly biased. We used *Egunkaria* as the main corpus, but we also used the EEBS corpus in some cases, as we will see below.

### 2.4 Words chosen

The criterion to choose the 40 words (15 nouns, 15 verbs and 10 adjectives) was that they should cover all possible combinations of frequency, polysemy and skew[1]. The first two can be objectively determined before starting to hand tag, but skew could only be determined by introspection. After choosing a word, the expected skew was sometimes different from the desired skew.

A secondary criterion was the overlap with the words in other languages, and the overlap with a number of verbs that are being used for subcategorization and diathesis alternation studies in our group.

The English task organizers and the Spanish task organizers provided us with half of the words chosen in their lexical-sample task. This information could be used for cross-language mapping of word senses. Regarding the overlap with verbs, we plan to explore the influence of word senses in subcategorization and diathesis alternations.

We chose the first set of 40 words that covered more or less all combinations of the above phenomena from the set of translations of the words in the other tasks. This was done blindly, without knowing which specific word was chosen.

This first set was used to extract the examples (cf. following section), and the hand-taggers started to tag them. Unfortunately, for a number of words, all examples in the corpus referred to a single word sense. We had not foreseen this situation and took two measures:
1) Search for occurrences in the secondary EEBS corpus.
2) If occurrences of new senses were not found, then the word was discarded and a replacement word was chosen.

In order to find the replacement, the hand-tagger that was doing the arbitration scanned the examples of a word with similar polysemy and frequency and decided whether it had occurrences of more than one sense.

### 2.5 Selection of examples from corpora

The minimum number of examples for each word according to the task specifications was calculated as follows:

$$N=75+15*senses+6mword$$

where senses does not include the *nuances* (cf. section 2.2) and *mword* is the number of multiword terms that included the target word.

The minimum number of examples per word was extracted at random from the *Egunkaria* corpus, plus a 10% buffer. As explained in the previous section, for some words occurring in a single sense in this corpus, additional examples were taken from the secondary *EEBS* corpus. In this case, all available examples from *EEBS* were used, plus the examples from *Egunkaria* to meet the minimum number of examples required.

The context included 5 sentences, with the sentence with the target word appearing in the middle. Links were kept to the source corpus, document, and to the newspaper section when applicable.

The occurrences were split at random in training set (two thirds of all occurrences) and test set.

---

[1] By skew in this context, we mean the dominance of one sense over the others. It is given as the percentage of occurrences of the most frequent sense over all the others.

## 3 Hand tagging

Three persons, graduate linguistics students, took part in the tagging. They are familiar with word senses, as they are involved in the development of the Basque WordNet and cleaning the TEI version of the *Euskal Hiztegia* dictionary. The following procedure was defined for each word:

- The three of them would meet, read the definitions and examples given in the dictionary and discuss the meaning of each word sense. They tried to agree the meaning differences among the word senses.
- Two taggers independently tagged all examples for the word. No communication was allowed while tagging the word.
- Multiple tags were allowed, as well as the following tags: B new sense or multiword term, U unassignable. Examples with these tags were removed from the final release.
- A program was used to compute agreement rate and output those occurrences where there was disagreement grouped by the senses assigned.
- The third tagger, the referee, reviewed the disagreements and decided which one was correct.

For the word *itzal* (shadow), the disagreement was specially high. The taggers decided that the definitions and examples were too confusing, and decided to replace it with another word.

Overall, the two taggers agreed 75% of the time. Some words attained an agreement rate above 95% (e.g. nouns *kanal* – channel – or *tentsio* –tension – ), but others like *herri* – town/people/nation – attained only 52% agreement.

All in all, 5284 occurrences of the 40 words were released. On average, one hand-tagger took 0.41 minutes per occurrence and the other 0.55 minutes. The referee took 0.22 minutes per entry, including selection of replacement words. Time for arbitration meeting is also included.

## 4 Participants and Results

Three different teams and four systems took part in the tagging: John Hopkins University (JHU), Basque Country University (BCU-EHU-dlist-all and BCU-EHU-dlist-best) and University of Maryland (UMD). The third team submitted the results later, out of the Senseval competition. The results for the fine-grain scoring are shown in table 1, including the Most Frequent Sense baseline (MFS). Assuming full coverage, JHU attains the best performance. BCU-EHU-dlist-best has the best precision, but only tags 57% of the occurrences.

| Prec. | Recall | Attempted | System |
|---|---|---|---|
| 0,849 | 0,483 | 56,9% | BCU-ehu-dlist-best |
| 0,757 | 0,757 | 100% | JHU |
| 0,732 | 0,732 | 100% | BCU-ehu-dlist-all |
| 0,703 | 0,703 | 100% | UMD |
| 0,648 | 0,648 | 100% | MFS |

**Table 1:** results of systems and MFS baseline. UMD submitted results after the deadline.

## 5 Discussion

These are the main issues that we think are interesting for further discussion.

**Dictionary used.** Before designing the task, we had to choose between two possible dictionaries: the Basque WordNet and the *Euskal Hiztegia* dictionary. Another alternative was to start the lexicographer's work afresh, defining the word senses as the tagging proceeded. We thought the printed dictionary would provide clear-cut sense distinctions that would allow the tagging to be easier. After the tagging, the hand-taggers complained that this was not the case. They think that the tagging would be much more satisfactory had they defined the word senses directly from the corpus.

In particular, they were not allowed to introduce new senses or multiword terms, and such examples were discarded.

**Corpus used.** There was a mismatch between the dictionary and the corpus: the corpus was linked to a specific genre, and this resulted in having some senses which were not included in the dictionary. Besides, many senses in the dictionary did not appear in our corpus, and some words had to be replaced. This caused the taggers some overwork, but did not influence the quality of the result.

**Hand-tagging is a very unpleasant task.** When asked about future editions, the hand taggers suggested the following: "please do get somebody else". We have to note that the hand taggers are used to repetitive tasks, such as building the Basque WordNet or cleaning-up the TEI version of *Euskal Hiztegia.*

**Inter-tagger agreement.** Part of the disagreement was caused by typos and mistakes. Nevertheless, we think that the low inter-tagger agreement (75%) was caused mainly by the procedure used to tag the occurrences. The taggers met and tried to understand the word senses, but the fact is that it was only after tagging a few occurrences that they started to really conceptualise the word senses and draw specific lines among one sense and the others. If both taggers had been allowed to meet (at least once) while they were tagging, they could have discussed and agreed on a common conceptualisation. The referee found that most of the times whole sets of examples were systematically tagged differently by each of the taggers, that is, each of the taggers had a different criterion about the word sense applicable to that set of examples. The referee then had to decide on the tag for those sets of examples.

**Systems performing as good as inter-tagger agreement.** Traditionally**,** inter-tagger agreement has been used as an upperbound for the performance of machines in cognitive tasks. We think that in this case, a system may perform better on the Basque WSD task than a human, in the sense that if the taggers were evaluated against the gold standard they would score lower that the systems. In fact, current systems, which are still under development for Basque, reach the same performance as humans. Are machines performing better than humans? We think that inter-tagger agreement, at least as derived from the procedure used in this exercise, is not a real upperbound, and that systems can easily perform better.

The gold standard reflects the conceptualization of one human, the referee, which does not have to agree with the conceptualization made by other persons (specially if these are done in isolation). People disagree whether in a certain occurrence this word sense or the other applies, i.e. they can disagree in the meaning of the word senses as defined in the dictionary. In fact, trying to achieve a common ground when reading the dictionary definitions sometimes produced heated debate in the meetings.

If the gold standard reflects a systematic conceptualization of a person, machine learning algorithms can learn to replicate these conceptualization (categorizations in this case), and achieve high degrees of agreement with the person behind the gold standard. This does not mean that the system is smarter than the human taggers, but rather that the system has no opinion on his own, and just imitates one of the persons.

**Error reduction similar to English task.** The best recall for Basque was 75% vs. 64% of the MFS baseline. In English the best system achieved 64% recall vs. 47% of the most frequent sense baseline (called commonest baseline in the official results). It is clear that the skew of the Basque words allowed for higher results. On the other hand, the error reduction for Basque was 29%, compared to 32% for English. This implies that systems could effectively learn from the data in both tasks.

**No use of domain tags, full documents.** No system used the extra information provided by the full documents or the domain tags.

## 6 Future work

First of all, we plan to explore the use of other procedures for the hand-tagging. We think that the data attained high levels of quality (which has been shown by the error reduction attained by the participating systems over the MFS baseline), but still we are not satisfied with the sense inventory used.

Further analysis of the results of the participating systems is also planned, as Kappa statistics and the performance of the combination of the systems.